# Predictors of Childhood Vaccination Uptake in England: An Explainable Machine Learning Analysis of Longitudinal Regional Data (2021–2024)

Amin Noroozi[a], Sidratul Muntaha Esha[a], and Mansoureh Ghari[b]

[a] School of Engineering, Computing & Mathematical Sciences, University of Wolverhampton, Wolverhampton, WV1 1LY, UK

[b] School of Pharmacy, University of Wolverhampton, Wolverhampton, WV1 1NA, UK

**Corresponding Author:**
Amin Noroozi
Room 148, Alan Turing Building, School of Engineering, Computing & Mathematical Sciences, University of Wolverhampton, Wolverhampton, WV1 1LY, UK
Email: a.noroozifakhabi@wlv.ac.uk

**Summary**

**Background**: Childhood vaccination is a cornerstone of public health, yet disparities in vaccination coverage persist across England. These disparities are shaped by complex interactions among various factors, including geographic, demographic, socioeconomic, and cultural (GDSC) factors. While previous studies have investigated these predictors, most rely on static, cross-sectional data and traditional statistical approaches that assess individual or limited sets of variables in isolation. Such methods may fall short in capturing the dynamic and multivariate nature of vaccine uptake.

**Methods**: We conducted a longitudinal machine learning analysis of childhood vaccination coverage across 150 districts in England from 2021 to 2024. Using vaccination data from NHS records, we applied hierarchical clustering to group districts by vaccination coverage into low- and high-coverage clusters, using two, three, and six clusters to capture varying levels of disparity. A CatBoost classifier was then trained to predict districts' vaccination clusters using their GDSC data. Finally, the SHapley Additive exPlanations (SHAP) method was used to identify and interpret the most important predictors of vaccination disparities.

**Findings**: The optimal clustering involved two clusters, for which the CatBoost classifier achieved high accuracies of 92.1, 90.6, and 86.3 in predicting districts' vaccination clusters for the years 2021-2022, 2022-2023, and 2023-2024, respectively. SHAP analysis revealed that geographic, cultural, and demographic variables, particularly rurality, English language proficiency, percentage of foreign-born residents, and ethnic composition, were the most influential predictors of vaccination coverage. Contrary to common assumptions, rural districts were significantly more likely to have higher vaccination rates. Additionally, districts with lower vaccination coverage had significantly higher populations whose first language was not English, who were born outside the UK, or who were from ethnic minority groups. Surprisingly, socioeconomic variables, such as deprivation and employment, had consistently lower importance, especially in 2023–2024.

**Interpretation**: Vaccination disparities in England are primarily driven by geographic, demographic, and cultural factors rather than socioeconomic ones. Machine learning with explainable outputs offers actionable insights for public health planning, particularly for targeting vulnerable communities.

**Funding**: This study did not use any external funding.

# INTRODUCTION

**Background**

Childhood vaccination is a cornerstone of public health, significantly reducing the incidence of infectious diseases and preventing outbreaks that can have severe consequences for children and communities alike (1). Vaccination programs are among the most effective interventions for disease prevention, yet disparities in vaccine coverage persist globally and within nations (2-5). In England, while overall childhood immunization rates have been relatively high, regional disparities have raised concerns about inequalities in vaccine uptake and the potential risks associated with suboptimal coverage (6, 7). These disparities are influenced by a complex interplay of geographic, demographic, socioeconomic, and cultural (GDSC) factors, making it crucial to analyze vaccination trends over time and identify the key predictors of vaccine uptake (8-10).

**Importance**

Understanding the determinants of vaccination coverage is fundamental for effective public health planning and policymaking. Several studies have highlighted key predictors of vaccine uptake, including parental education (11), household income (12, 13), healthcare accessibility (14), and community trust in medical institutions (15, 16). Socioeconomic factors, in particular, have been shown to play a critical role, as lower-income households often face greater barriers to healthcare access, contributing to disparities in immunization rates (13, 17). Moreover, vaccine hesitancy, driven by misinformation, cultural beliefs, and concerns about vaccine safety, has emerged as a significant challenge in maintaining high vaccination rates (18). Geographic disparities in vaccine coverage add another layer of complexity to immunization programs. For example, some studies have reported lower vaccination rates in rural areas, primarily due to limited access to healthcare services, while urban centres often exhibit heterogeneous vaccination patterns, influenced by cultural and socioeconomic diversity (19, 20). These disparities highlight the need for targeted interventions and region-specific strategies to improve immunization coverage and reduce public health risks.

Current research predominantly relies on static, cross-sectional data, which restricts our ability to capture the dynamic nature of vaccination trends over time. Additionally, most studies focus on individual predictors or diseases, failing to provide a comprehensive comparison of the relative importance of various factors influencing different diseases' vaccination coverage. To overcome these limitations, a holistic, data-driven approach is essential for integrating multiple determinants and diseases and supporting targeted public health interventions.

Machine learning has emerged as a powerful tool in epidemiological research, enabling researchers to uncover hidden patterns in vaccination trends (21-23). Machine learning methods can process large-scale high-dimensional data more accurately than traditional statistical techniques, making them reliable options for addressing the problems mentioned above in vaccination research (22, 24). However, despite its potential, the application of machine learning to longitudinal vaccination coverage trends remains underexplored.

**Goals of This Study**

This study aims to bridge this gap by investigating childhood vaccination disparities across a broad range of diseases in England between 2021 and 2024, using a machine learning framework. To achieve this, we first apply hierarchical clustering to group districts based on their vaccination coverage levels and analyse how these clusters evolve over time. We then train a CatBoost (CB) classifier to predict districts' vaccination clusters using their GDSC factors. Finally, we employ SHapley Additive exPlanations (SHAP) (25) to interpret the model and identify the most influential predictors contributing to vaccination coverage disparities across England.

## METHODS

### Study Setting and Data

Two types of data from three consecutive years, 2021–2022, 2022–2023, and 2023–2024, were collected for this study. Each study year runs from 1 April to 31 March of the following year, in alignment with the NHS national statistics format (26), which is published each September. The first dataset includes 14 types of vaccination data sourced from the NHS Immunization Statistics (26-29), providing annual records of immunization rates at the Upper-Tier Local Authority (UTLA) level for 150 districts across England with a focus on early vaccinations for children under 5 years old. The second dataset used in this study contains GDSC variables, grouped into nine categories. For simplicity, we refer to these nine categories collectively as GDSC data throughout this paper. The data were collected from statistics provided by the UK government, including the English Indices of Multiple Deprivation (IMD) (30), census data (31), and Rural-Urban Classification (32). Table 1 shows the abbreviations used for the data in the first and second datasets in this study and the corresponding descriptions.

As can be seen from Table 1, coverage rates of 14 vaccinations have been included in our analysis to ensure the comprehensiveness of the analysis, concentrating on early childhood immunizations that are part of the UK's routine immunization schedule. These vaccinations cover 13 unique diseases, including Diphtheria, Tetanus, Pertussis (Whooping Cough), Polio, Haemophilus influenzae type B (Hib), Hepatitis B, Meningococcal Disease (Group B), Meningococcal Disease (Group C), Measles, Mumps, Rubella, Pneumococcal Disease, and Rotavirus Disease.

For the second dataset, "IMD-Average score", "IMD-Proportion deprived", "Long-term unemployed", "Routine occupations", and "No qualifications" primarily represent the socioeconomic condition of the population. The variables "Born Outside the UK" and "Ethnic Minority" can be classified as demographic factors, while "Rurality" captures the geographic characteristics of each district. Additionally, both "Ethnic Minority" and "English Proficiency" can be considered cultural variables, as research indicates that cultural elements associated with different ethnic groups, such as beliefs, traditions, and levels of trust in healthcare systems, can significantly affect attitudes toward vaccination (33). Moreover, since all GDSC factors are calculated separately for each district across England, the impact of geography is inherently included in their calculations, which in turn, impacts the vaccination disparities in respective regions (34).

### Machine Learning Analysis

The machine learning analysis in this study consisted of three steps, outlined as follows:

**Step 1**: Districts were clustered into two, three, and six clusters based on their vaccination rates extracted from the first dataset. In this step, a cluster label was assigned to each district, which served as the target variable for the next step. The assigned cluster reflects the district's level of vaccination coverage. For example, if the districts are grouped into two clusters, one cluster would represent districts with high vaccination rates and the other those with low vaccination rates.

**Step 2**: In the second step, a CB classifier was employed to classify districts into their respective vaccination clusters using GDSC variables from the second dataset. This step is equivalent to predicting the cluster labels using the GDSC data, rather than the vaccination rates themselves.

**Step 3**: In this step, we used SHAP values derived from the CB model to evaluate the contribution of each of the nine GDSC factors in predicting the cluster labels assigned to districts in Step 2.

Clustering the districts in Step 1 using conventional methods is challenging, particularly when considering all vaccination rates from the first dataset (Table 1). Additionally, the optimal number of clusters is a hyperparameter that must be defined prior to running the clustering algorithm. To address these challenges, we employed a hierarchical clustering method, with the optimal number of clusters determined using a dendrogram (35). Hierarchical clustering is a method that creates a hierarchy of clusters in a tree-like structure known as a dendrogram. This approach is particularly useful in clinical research involving heterogeneous data, as it allows for the exploration of similarities between observations and clusters, which can be visualized using dendrograms (35).

To classify data in step 2, we evaluated various methods, with the CB classifier demonstrating superior performance; consequently, it was selected for this research. Tree-based classifiers such as CB are renowned for their robustness and accuracy in classification tasks across diverse datasets (21, 36, 37). The CB algorithm, when combined with SHAP, provides a robust and interpretable approach for evaluating feature importance. This method enables the identification of the most influential predictors in the classification process by quantifying each feature's contribution to the model's predictions. (25, 38). This capability allows us to determine the key predictors influencing vaccination coverage rates across districts. Accordingly, we used this mechanism to extract feature importance values from the CB classifier and assess the contribution of the nine GDSC factors outlined in Table 1.

To ensure the robustness of our results, we implemented a five-fold cross-validation (CV) approach, which is a standard technique to enhance model generalizability and prevent overfitting (39–43). In this method, the dataset is initially partitioned into five equal folds. In each iteration, four folds are utilized for training, while the remaining fold serves for testing. This process is repeated five times, ensuring that each fold serves as the test set once. The final evaluation metrics and feature importance values were computed by averaging the results across all folds, ensuring a comprehensive assessment of the model's performance and the relevance of each feature.

**Table 1.** The abbreviations and descriptions of data used in this study.

| | Abbreviation | Description |
|---|---|---|
| **First dataset (vaccination rates)** | DTaP_IPV_5y | 4 in 1 vaccine (Diphtheria, tetanus, pertussis and polio) at 5 years |
| | DTaP_IPV_Hib_5y | 5-in-1 vaccine (diphtheria, pertussis, tetanus, polio, Haemophilus influenzae type b) at 5 years |
| | DTaP_IPV_Hib_HepB_12m | 6-in-1 vaccine (diphtheria, pertussis, tetanus, polio, Haemophilus influenzae type b, hepatitis B) at 12 months |
| | DTaP_IPV_Hib_HepB_24m | 6-in-1 vaccine (diphtheria, pertussis, tetanus, polio, by Haemophilus influenzae type b, hepatitis B) at 24 months |
| | Hib_MenC_24m | Haemophilus Influenzae type b and meningococcal group C (Hib/MenC) at 24 months |
| | Hib_MenC_5y | Haemophilus Influenzae type b and meningococcal group C (Hib/MenC) at 5 years |
| | MenB_12m | Meningococcal disease (group b) at 12 months |
| | MenB_booster_24m | Meningococcal disease (group b) at 24 months |
| | MMR_24m | Measles mumps rubella (MMR) (1st dose at 24 months) |
| | MMR1_5y | Measles mumps rubella (MMR) (1st dose at 5 years) |
| | MMR2_5y | Measles mumps rubella (MMR) (2nd dose at 5 years) |
| | PCV_12m | Pneumococcal disease at 12 months |
| | PCV_24m | Pneumococcal disease at 24 months |
| | Rota_12m | Rotavirus at 12 months |
| **Second dataset (GDSC data)** | IMD-Average score | Represents the mean deprivation score across all Lower Layer Super Output Areas (LSOAs) within a given UTLA |
| | IMD-Proportion deprived | Refers to the percentage of LSOAs within a UTLA that fall into the most deprived 10% of all LSOAs in England |
| | Long-term unemployed | Percentage of individuals classified as long-term unemployed, derived from Census data |
| | Routine occupations | Percentage of people in routine occupation |
| | No qualifications | Percentage of people with no qualification |
| | English proficiency | Percentage of individuals whose primary language is not English |
| | Ethnic minority | Representation of ethnic minority populations within each UTLA |
| | Born outside UK | Percentage of residents born outside the UK |
| | Rurality | Categorical variable with 6 categories showing the proportion of the population living in rural areas, ranging from Urban with Major Conurbation (category 1) to Mainly Rural (category 6) |

## RESULTS

### Districts' Longitudinal Vaccination Clusters

To identify patterns in vaccination coverage rates across districts, we initially employed hierarchical clustering and examined the resulting dendrograms to determine the optimal number of clusters for each study year: 2021–2022, 2022–2023, and 2023–2024. In all three years, the dendrograms indicated that two clusters provided the best fit. However, to capture finer-grained disparities, we also conducted supplementary clustering using three and six clusters, offering a more detailed stratification of vaccination coverage levels. An example of the dendrogram used to derive the optimal number of clusters for the year 2023–2024 is provided in Figure W1 of the supplementary web appendix.

Figure 1 illustrates disparities in childhood vaccination coverage across England, using two, three, and six clusters for the years 2021–2022, 2022–2023, and 2023–2024. Table 2 presents the average vaccination rates for districts within each cluster. The complete list of districts grouped by cluster for each year can be found in Tables W1–W9 in the supplementary web appendix.

As shown in Figure 1(a), the overall pattern of vaccination disparities remained largely consistent between 2021–2022 and 2022–2023. However, a substantial shift is evident in 2023–2024, where district clustering patterns diverged markedly from the previous two years. In particular, several large and densely populated districts, including Birmingham, Manchester, Liverpool, Barnet, and Croydon, transitioned from the low-vaccination coverage cluster in 2021–2022 and 2022–2023 to the high coverage cluster in 2023–2024, indicating notable improvements in vaccine uptake. Conversely, two of England's largest districts, Cambridgeshire and Cumbria, shifted from high to low vaccination coverage clusters in 2023–2024, suggesting a decline in performance. A more granular clustering with six groups, shown in Figure 1(c) and detailed in Tables W3, W6, and W9, reveals that Hackney consistently had the lowest vaccination coverage in the first two years. However, in 2023–2024, Haringey and Lambeth replaced Hackney in the lowest coverage cluster. Despite these changes, all three districts have consistently been among the lowest-performing areas in terms of childhood vaccination rates.

**Classification**

After clustering the districts based on their vaccination coverage and assigning cluster labels (as shown in Figure 1 and detailed in Tables W1–W9), we employed the CB classifier to predict these labels using the GDSC variables outlined in Table 1. Table 3 presents the classification performance metrics, including accuracy, precision, recall, and F1 score, across different clustering scenarios (two, three, and six clusters) and study years. These metrics were averaged across all classes and cross-validation folds. For detailed definitions of these evaluation metrics, please refer to (44).

Across all three years, the CB classifier demonstrated high predictive performance using two clusters, with accuracy consistently exceeding 86%. The highest accuracy was achieved in 2021–2022, reaching 92.1%. As expected, the classification accuracy declined with an increasing number of clusters, particularly when using six clusters. Nonetheless, performance remained acceptable under the three-cluster scenario. These results are consistent with the dendrogram analysis (Figure W1), which identified two as the optimal number of clusters based on hierarchical clustering.

**The Most Important Predictors**

We used the trained CB classifiers to compute SHAP values, which are interpreted in this section as measures of feature importance. Figure 2 presents the feature importance results for the two-cluster configuration in the years 2021–2022 and 2022–2023. Results for the three- and six-cluster models are provided in Figure W2 in the web appendix. As illustrated, the features "Rurality," "Born Outside UK," "English Proficiency," and "Ethnic Minority" consistently emerged as the most influential predictors across all years and clustering configurations.

Although the three demographic and cultural features, Born Outside UK, Ethnic Minority, and English Proficiency, may have relatively high intercorrelation, we retained all of them in our models. Empirical testing revealed that removing any one of these features resulted in a notable decline in model performance, suggesting that each contributes distinct and complementary information. One plausible explanation is that a significant portion of individuals may belong to an ethnic minority group or be born outside the UK but still speak English as their first language. The SHAP analysis supports this hypothesis, as all three features consistently demonstrated high importance scores, indicating they each capture unique aspects relevant to predicting vaccination coverage disparities.

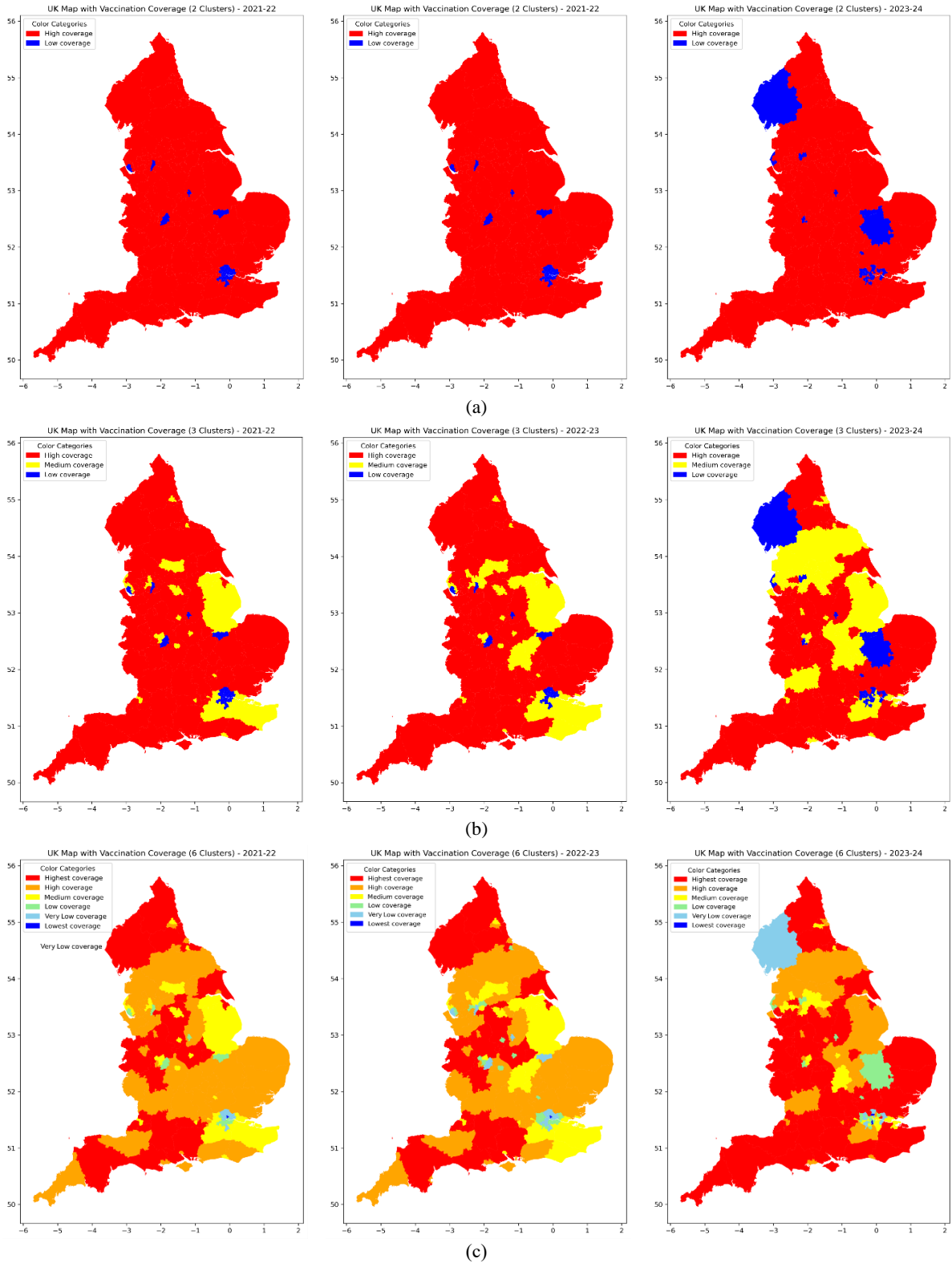

**Figure 1**. Vaccination coverage rate maps for England using (a) two clusters, (b) three clusters, and (c) six clusters, from left to right, for the years 2021-2022, 2022-2023, and 2023-2024, respectively.

**Table 2.** Average vaccination rates for different diseases across England, grouped by two, three, and six clusters for the years 2021–2022, 2022–2023, and 2023–2024.

| Year | Number of clusters | Cluster label | DTaP_IPV_5y | DTaP_IPV_Hib_5y | DTaP_IPV_Hib_HepB_12m | DTaP_IPV_Hib_HepB_24m | Hib_MenC_24m | Hib_MenC_5y | MenB_12m | MenB_booster_24m | MMR_24m | MMR1_5y | MMR2_5y | PCV_12m | PCV_24m | Rota_12m |
|---|---|---|---|---|---|---|---|---|---|---|---|---|---|---|---|---|
| 2021-2022 | 2 | L | 69.9 | 89.2 | 85.2 | 86.3 | 78.8 | 84.6 | 84.8 | 76.8 | 79.0 | 87.1 | 72.6 | 87.6 | 79.4 | 82.9 |
| | | H | 87.3 | 95.5 | 93.4 | 94.5 | 91.3 | 93.3 | 93.1 | 90.4 | 91.5 | 94.8 | 88.51 | 95.2 | 91.7 | 91.5 |
| | 3 | L | 69.9 | 89.3 | 85.2 | 86.3 | 78.8 | 84.6 | 84.8 | 76.8 | 79.0 | 87.1 | 72.6 | 87.7 | 79.4 | 82.9 |
| | | M | 81.8 | 93.6 | 90.8 | 92.0 | 87.1 | 90.3 | 90.3 | 85.7 | 87.3 | 92.6 | 83.2 | 93.2 | 87.8 | 88.5 |
| | | H | 89.5 | 96.3 | 94.4 | 95.5 | 93.0 | 94.5 | 94.3 | 92.4 | 93.2 | 95.7 | 90.7 | 96.0 | 93.3 | 92.8 |
| | 6 | Ls | 56.1 | 82.4 | 64.0 | 70.6 | 61.6 | 78.2 | 64.5 | 60.2 | 65.4 | 83.5 | 58.9 | 70.9 | 64.3 | 61.7 |
| | | VL | 66.6 | 88.7 | 84.4 | 85.0 | 76.3 | 83.1 | 83.8 | 74.3 | 76.6 | 85.8 | 68.5 | 87.2 | 77.0 | 82.4 |
| | | L | 73.4 | 90.2 | 87.3 | 88.5 | 81.9 | 86.2 | 86.9 | 79.8 | 81.8 | 88.3 | 76.9 | 89.1 | 82.3 | 84.8 |
| | | M | 81.8 | 93.6 | 90.8 | 92.0 | 87.1 | 90.3 | 90.3 | 85.7 | 87.3 | 92.6 | 83.2 | 93.2 | 87.8 | 88.5 |
| | | H | 88.4 | 95.9 | 93.6 | 94.9 | 91.9 | 93.8 | 93.4 | 91.2 | 92.1 | 95.2 | 89.5 | 95.4 | 92.2 | 91.8 |
| | | Hst | 91.5 | 97.1 | 95.9 | 96.6 | 94.9 | 95.6 | 95.7 | 94.4 | 95.0 | 96.6 | 92.6 | 97.0 | 95.2 | 94.5 |
| 2022-2023 | 2 | L | 68.8 | 87.4 | 85.9 | 86.2 | 79.0 | 82.4 | 84.5 | 77.4 | 80.2 | 85.4 | 70.2 | 87.9 | 79.3 | 82.1 |
| | | H | 85.6 | 94.1 | 92.9 | 93.7 | 90.4 | 91.7 | 92.2 | 89.3 | 90.9 | 93.7 | 86.8 | 94.7 | 90.1 | 89.9 |
| | 3 | L | 68.8 | 87.4 | 85.9 | 86.2 | 79.0 | 82.4 | 84.5 | 77.4 | 80.2 | 85.4 | 70.2 | 87.9 | 79.3 | 82.1 |
| | | M | 81.3 | 92.0 | 90.9 | 91.6 | 87.3 | 88.9 | 89.8 | 85.6 | 88.0 | 91.5 | 82.7 | 93.0 | 86.5 | 87.2 |
| | | H | 88.8 | 95.6 | 94.4 | 95.3 | 92.7 | 93.7 | 93.9 | 92.1 | 93.0 | 95.2 | 89.8 | 95.9 | 92.8 | 91.8 |
| | 6 | Ls | 62.2 | 81.1 | 67.8 | 77.7 | 66.7 | 74.6 | 68.5 | 64.0 | 69.5 | 80.2 | 62.7 | 75.0 | 68.3 | 64.9 |
| | | VL | 66.4 | 84.8 | 84.1 | 86.2 | 77.8 | 79.0 | 83.3 | 75.8 | 78.2 | 81.9 | 66.2 | 87.0 | 77.3 | 81.6 |
| | | L | 73.2 | 87.6 | 87.3 | 88.2 | 81.8 | 83.1 | 86.3 | 80.1 | 82.2 | 86.4 | 74.3 | 88.9 | 80.9 | 83.6 |
| | | M | 78.9 | 91.2 | 88.9 | 90.6 | 85.5 | 87.5 | 88.1 | 83.7 | 86.1 | 90.2 | 80.5 | 91.6 | 84.7 | 85.6 |
| | | H | 83.7 | 93.0 | 91.9 | 92.9 | 89.8 | 89.5 | 91.3 | 88.2 | 90.0 | 92.5 | 85.2 | 94.0 | 89.2 | 89.1 |
| | | Hst | 88.7 | 95.5 | 94.6 | 95.4 | 93.0 | 93.5 | 94.2 | 92.2 | 93.2 | 95.2 | 89.7 | 96.0 | 92.8 | 92.3 |
| 2023-2024 | 2 | L | 70.1 | 86.2 | 84.8 | 86.8 | 79.3 | 81.1 | 84.0 | 77.5 | 79.9 | 84.4 | 70.8 | 87.2 | 78.8 | 81.5 |
| | | H | 85.2 | 93.9 | 92.6 | 93.7 | 90.5 | 91.1 | 92.1 | 89.3 | 90.8 | 93.4 | 86.5 | 94.5 | 90.1 | 89.9 |
| | 3 | L | 70.1 | 86.2 | 84.8 | 86.8 | 79.3 | 81.1 | 84.0 | 77.5 | 79.9 | 84.4 | 70.8 | 87.3 | 78.8 | 81.5 |
| | | M | 81.4 | 92.1 | 90.5 | 91.8 | 87.7 | 88.5 | 89.7 | 86.0 | 88.1 | 91.4 | 82.9 | 92.8 | 87.0 | 87.4 |
| | | H | 88.7 | 95.5 | 94.6 | 95.4 | 93.0 | 93.5 | 94.2 | 92.2 | 93.2 | 95.2 | 89.7 | 96.0 | 92.8 | 92.3 |
| | 6 | Ls | 56.1 | 82.4 | 64.0 | 70.6 | 61.6 | 78.2 | 64.5 | 60.2 | 65.4 | 83.5 | 58.9 | 70.9 | 64.3 | 61.7 |
| | | VL | 66.6 | 88.7 | 84.4 | 85.0 | 76.3 | 83.1 | 83.8 | 74.3 | 76.6 | 85.8 | 68.5 | 87.2 | 77.0 | 82.4 |
| | | L | 73.4 | 90.2 | 87.3 | 88.5 | 81.9 | 86.2 | 86.9 | 79.8 | 81.8 | 88.3 | 76.9 | 89.1 | 82.3 | 84.8 |
| | | M | 81.8 | 93.6 | 90.8 | 92.0 | 87.1 | 90.3 | 90.3 | 85.7 | 87.3 | 92.6 | 83.2 | 93.2 | 87.8 | 88.5 |
| | | H | 88.4 | 95.9 | 93.6 | 94.9 | 91.9 | 93.8 | 93.4 | 91.2 | 92.1 | 95.2 | 89.5 | 95.4 | 92.2 | 91.8 |
| | | Hst | 91.5 | 97.1 | 95.9 | 96.6 | 94.9 | 95.6 | 95.7 | 94.4 | 95.0 | 96.6 | 92.6 | 97.0 | 95.2 | 94.5 |

Ls: Lowest, VL: Very Low, L: Low, M: Medium, H: High, Hst: Highest

**Table 3.** Classification performance metrics (%), accuracy, precision, recall, and F1 score for predicting vaccination coverage clusters in England using GDSC data.

| Year | Metric | 2 cluster | 3 cluster | 6 cluster |
|---|---|---|---|---|
| 2021-2022 | Accuracy | 92.1 | 77.3 | 48.7 |
| | Precision | 88.9 | 72.5 | 42.9 |
| | Recall | 86.5 | 71.9 | 45.5 |
| | F1 score | 86.2 | 71.4 | 42.1 |
| 2022-2023 | Accuracy | 90.6 | 76.0 | 42.7 |
| | Precision | 84.3 | 76.1 | 42.3 |
| | Recall | 77.0 | 71.8 | 41.8 |
| | F1 score | 77.9 | 72.1 | 40.6 |
| 2023-2024 | Accuracy | 86.3 | 64.6 | 49.3 |
| | Precision | 80.5 | 61.7 | 33.4 |
| | Recall | 70.0 | 59.8 | 32.8 |
| | F1 score | 70.4 | 59.9 | 31.6 |

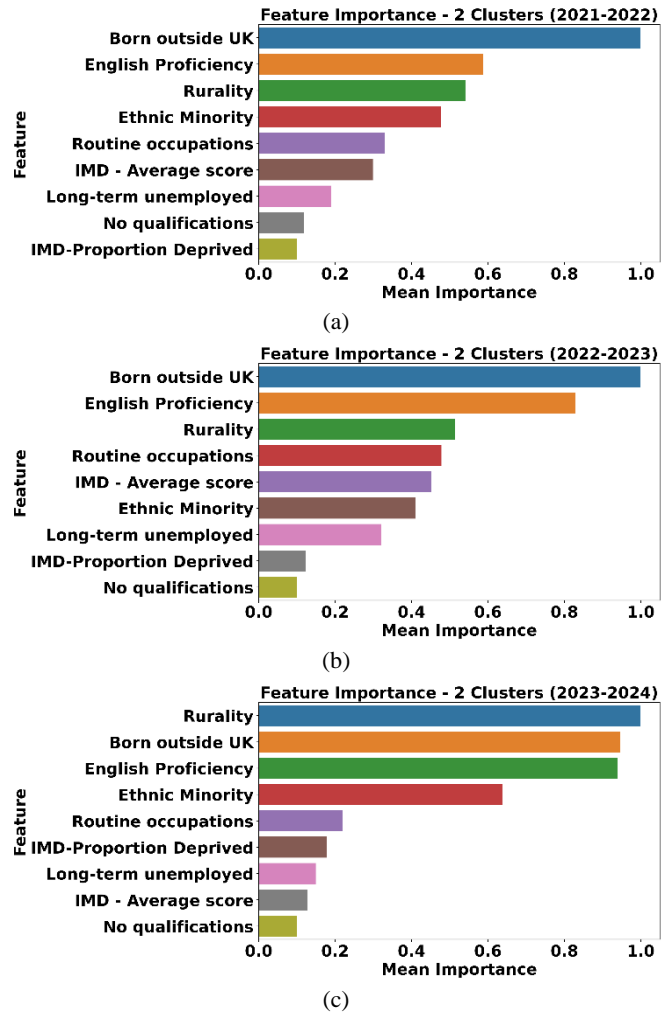

**Figure 2**. Most important predictors of vaccination coverage rates in England, based on SHAP values from the two-cluster model for (a) 2021–2022, (b) 2022–2023, and (c) 2023–2024.

Although Rurality emerged as one of the most important determinants in the SHAP analysis, this does not imply that rural districts exhibit lower vaccination coverage. In fact, the data reveal the opposite trend. As shown in Tables W1, W4, and W7, only 2, 2, and 3 districts, respectively, within the low-coverage cluster had a rurality classification greater than 1. According to Table 1, a rurality level of 1 corresponds to urban areas, indicating that the majority of low-coverage districts are, in fact, highly urbanized.

To explore this relationship further, we examined the distribution of rurality levels across vaccination clusters for the two-cluster model in all study years, as depicted in Figure W3 of the supplementary web appendix. The visualizations confirm that most districts with a rurality level above 1, i.e., more rural areas, are clustered in the high-vaccination coverage group. However, it is important to note that rurality alone is not a sufficient predictor of vaccination coverage. A considerable number of urban districts (rurality level 1) appear in both high and low coverage clusters, suggesting that other factors are influencing coverage disparities within urban settings.

To better understand the distribution patterns of the four most influential features, we present their box plots by vaccination coverage cluster in Figures W4–W7 of the supplementary web appendix. These visualizations show that low-coverage districts tend to be highly urbanized and have higher proportions of residents whose primary language is not English, who were born outside the UK, or who belong to ethnic minority groups. To support these observations, we conducted hypothesis testing comparing the distributions of these features across the two vaccination coverage clusters. The results indicated that the differences were statistically significant ($p < 0.05$) for all four features.

The socioeconomic factors presented in Figure 2 exhibit substantially lower SHAP importance values compared to the top four predictors, which are predominantly geographic, demographic, and cultural. With the exception of the two-cluster model for the year 2022–2023, where two socioeconomic variables, "Routine Occupation" and "IMD-Average Score", showed importance values comparable to the leading features, the classification in all other years and clustering configurations was consistently dominated by Rurality, Born Outside UK, English Proficiency, and Ethnic Minority. This trend is particularly pronounced in the 2023–2024 model, where the relative contribution of socioeconomic variables is markedly lower than in previous years. These findings suggest that while socioeconomic conditions remain relevant, they are less predictive of vaccination coverage disparities compared to geographic, demographic, and cultural characteristics, especially in the most recent year of analysis.

The relatively low SHAP importance values associated with socioeconomic factors, as shown in Figure 2, suggest that these variables do not play a primary role in explaining the disparities in vaccination rates between rural and urban districts. Instead, the feature importance results indicate that the observed differences are predominantly driven by demographic and cultural factors, specifically, variations in the proportion of residents born outside the UK, levels of English proficiency, and ethnic composition between rural and urban areas.

**LIMITATIONS**

The use of UTLA-level aggregated data in this study may restrict granularity and potentially mask important within-district variation. While the GDSC data are useful for national-level analyses, several studies suggest that individual-level predictors, such as parental attitudes, household composition, and vaccine confidence, may influence vaccination decisions (11, 15). Although some of these influences may be indirectly captured through demographic or socioeconomic proxies within the GDSC data, their effects are not explicitly modeled. Future research employing longitudinal, individual-level datasets would enhance the interpretability of predictive models and offer a more nuanced understanding of the behavioural and contextual drivers behind vaccination uptake.

This study did not include healthcare system variables such as GP density, service accessibility, or delivery logistics, which previous studies have found to be important (14, 34). While some of these factors may be indirectly reflected in certain GDSC variables, such as IMD, their individual contributions were not explicitly modeled. Future research could explore the relative impact of healthcare infrastructure variables alongside demographic and socioeconomic factors. Additionally, vaccine-specific concerns, such as hesitancy surrounding MMR or newer vaccines like MenB, were not captured in this study due to the absence of attitudinal data.

Finally, although model performance was validated through cross-validation, and SHAP analysis confirmed the relevance of retained features, the presence of correlated variables (e.g., ethnic minority status and language proficiency) may introduce redundancy and complicate interpretation. However, as demonstrated, removing any of these features significantly reduced model accuracy, suggesting each provides unique information.

Despite these limitations, the study provides a robust foundation for targeted policy and intervention design, and it highlights the utility of explainable machine learning in public health surveillance.

**DISCUSSION**

This study provides a comprehensive longitudinal analysis of childhood vaccination coverage disparities across England from 2021 to 2024, using an interpretable machine learning approach. By leveraging SHAP values within a CB classifier, we identified that the most influential predictors of vaccine coverage disparities were consistently geographic, demographic, and cultural variables, specifically rurality, English proficiency, proportion of ethnic minorities, and foreign-born population, rather than traditional socioeconomic indicators. We also observed that districts with lower vaccination rates tended to be more urban and had higher proportions of ethnic minority residents, foreign-born populations, and individuals whose first language was not English.

These findings align with prior research highlighting the significance of cultural and language barriers in vaccine uptake. For instance, one study (9) noted that vaccine confidence varied substantially across demographic groups, with minority and immigrant populations exhibiting greater hesitancy. Similarly, another study (18) emphasized the role of trust in healthcare systems and cultural beliefs as key contributors to vaccine acceptance or refusal. Our study adds to this literature by demonstrating, through quantitative feature importance, that these variables are more predictive of district-level disparities in England than socioeconomic factors like deprivation or employment status.

Our findings highlight the consistent and significant influence of ethnic minority representation on childhood vaccination coverage across all study years, as reflected in the SHAP importance values. Furthermore, as illustrated in Figure W7 of the supplementary web appendix, districts with higher proportions of ethnic minority populations were more likely to fall into the lower vaccination coverage clusters. This is consistent with previous research demonstrating that ethnic minority communities often face systemic barriers to vaccination, including reduced access to culturally appropriate health information, language barriers, and historical mistrust in healthcare systems (9, 33). Additionally, vaccine hesitancy may be more prevalent in some ethnic groups due to circulating misinformation, perceived discrimination, or negative healthcare experiences, which can undermine trust and reduce uptake (18). While the ethnic minority variable was correlated with other cultural and demographic features, such as English proficiency and foreign-born status, our model showed that each provided unique predictive value. This suggests that ethnic identity captures specific dimensions of community experience and cultural context that are not fully explained by language or migration status alone.

The association between English proficiency and vaccination uptake is especially noteworthy. Previous literature has emphasized that limited proficiency in the dominant language of a healthcare system can impair access to health information, reduce trust, and discourage vaccine-seeking behavior (11, 15). This highlights the importance of culturally competent communication and multilingual public health messaging in enhancing vaccine coverage among diverse communities.

Interestingly, our analysis found that rural districts generally had higher vaccination coverage than their urban counterparts, contrary to common assumptions that rurality correlates with reduced healthcare access (14, 45). This suggests that in the English context, urban complexity, diversity, and healthcare engagement dynamics may be more salient in shaping disparities than physical distance to services. This also mirrors the findings of (20), who reported heterogeneity within urban settings, particularly among low-income, diverse communities.

Furthermore, socioeconomic variables such as deprivation scores, unemployment rates, and educational attainment, which have historically been linked to health disparities, ranked lower in importance in our predictive models. This contrasts with earlier studies (8, 17) that identified a clear inverse relationship between deprivation and vaccine uptake. However, our findings may suggest a contextual shift in England, where universal access to NHS services has mitigated the direct impact of material deprivation on vaccination behaviour, while geographic, demographic, and cultural barriers have become more salient. Additionally, these results may reflect the effectiveness of policy interventions aimed at reducing access-related barriers for economically disadvantaged populations. This may suggest a shift in the relative influence of structural versus cultural predictors in contemporary England, possibly driven by widespread access to NHS vaccination services and evolving patterns of community engagement.

Notably, districts such as Birmingham, Manchester, and Liverpool showed marked improvement in vaccine coverage over the study period. These gains could reflect the successful implementation of targeted interventions, local outreach, or the impact of national campaigns. In contrast, formerly high-performing districts such as Cambridgeshire and Cumbria experienced declining coverage, echoing concerns raised by (6) about fluctuating local uptake despite overall national efforts.

By integrating machine learning with interpretable outputs, this study offers actionable insights for policymakers and public health practitioners. The findings support the need for localized, culturally informed strategies that address language proficiency, ethnic representation, and community trust as key levers for improving childhood immunization coverage. For example, multilingual outreach, collaboration with trusted local leaders, and culturally sensitive education campaigns could address low coverage in ethnically diverse urban areas. Additionally, the observed variation over time supports the need for adaptive strategies that respond to changing local conditions rather than static national policies.

**Contributors**
Amin Noroozi conceived and designed the study, developed the methodology, supervised the analysis, validated the results, and wrote the original draft. Sidratul Muntaha Esha contributed to data curation, formal analysis, project administration, validation, and visualization. Mansoureh Ghari contributed to the investigation, formal analysis, and manuscript review and editing.

**Declaration of interests**
No competing interests declared.

**Data sharing**
Upon acceptance, the data and code used in this study will be made available via Dr Amin Noroozi's GitHub repository: https://github.com/AminNoroozi/

**Acknowledgments**
The authors used ChatGPT (OpenAI, GPT-4) to assist with language editing and proofreading. Prompts were limited to requests for clarity, grammar correction, and academic tone refinement. All content was reviewed and verified by the authors

# "Supplementary Web Appendix"

**Hierarchical Clustering Dendrogram**

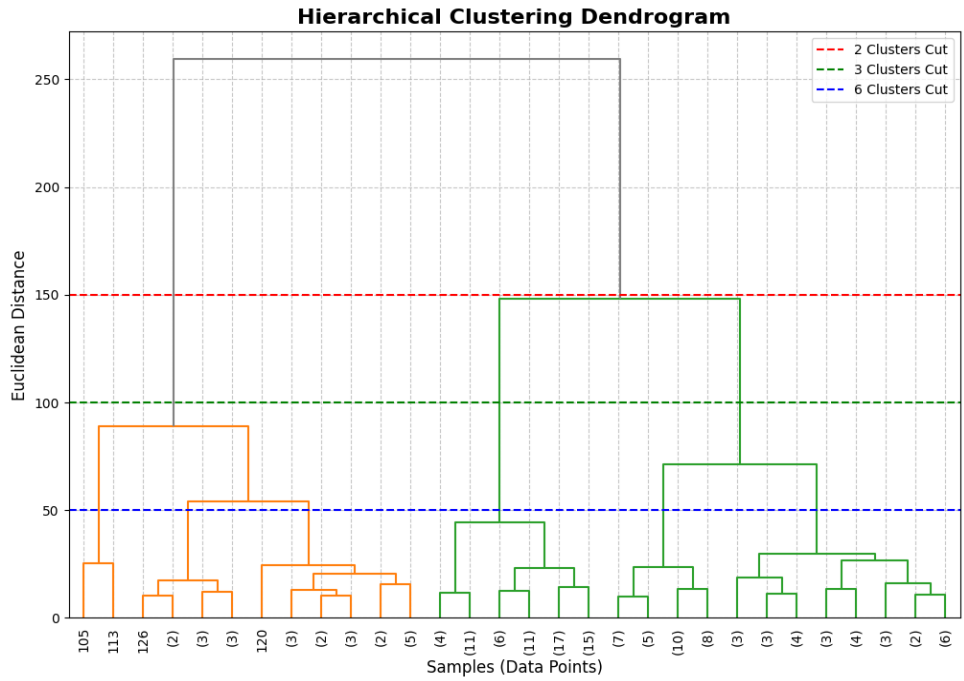

**Figure W3**. Hierarchical clustering dendrogram for vaccination rates in England in 2023-2024.

# Districts List by Vaccination Cluster

**Table W4**. List of districts in different vaccination coverage clusters for 2021-22 using two clusters

| High | | | | | Low |
|---|---|---|---|---|---|
| Hartlepool | Thurrock | Bury | Kirklees | Nottinghamshire | Nottingham |
| Middlesbrough | Medway | Oldham | Leeds | Oxfordshire | Peterborough |
| Redcar and Cleveland | Bracknell Forest | Rochdale | Wakefield | Somerset | Manchester |
| Stockton-on-Tees | West Berkshire | Salford | Gateshead | Staffordshire | Liverpool |
| Darlington | Reading | Stockport | Bexley | Suffolk | Birmingham |
| Halton | Slough | Tameside | Bromley | Surrey | Barking and Dagenham |
| Warrington | Windsor and Maidenhead | Trafford | Ealing | Warwickshire | Barnet |
| Blackburn with Darwen | Wokingham | Wigan | Harrow | West Sussex | Brent |
| Blackpool | Milton Keynes | Knowsley | Havering | Worcestershire | Camden |
| Kingston upon Hull | Brighton and Hove | St Helens | Hillingdon | Rutland | Croydon |
| East Riding of Yorkshire | Portsmouth | Sefton | Hounslow | | Enfield |
| North East Lincolnshire | Southampton | Wirral | Kingston upon Thames | | Greenwich |
| North Lincolnshire | Isle of Wight | Barnsley | Sutton | | Hackney |
| York | County Durham | Doncaster | Cambridgeshire | | Hammersmith and Fulham |
| Derby | Cheshire East | Rotherham | Cumbria | | Haringey |
| Leicester | Cheshire West and Chester | Sheffield | Derbyshire | | Islington |
| Herefordshire | Shropshire | Newcastle upon Tyne | Devon | | Kensington and Chelsea |
| Telford and Wrekin | Cornwall | North Tyneside | East Sussex | | Lambeth |
| Stoke-on-Trent | Wiltshire | South Tyneside | Essex | | Lewisham |
| Bath and North East Somerset | Bedford | Sunderland | Gloucestershire | | Merton |
| Bristol | Central Bedfordshire | Coventry | Hampshire | | Newham |
| North Somerset | Northumberland | Dudley | Hertfordshire | | Redbridge |
| South Gloucestershire | Bournemouth, Christchurch and Poole | Sandwell | Kent | | Richmond upon Thames |
| Plymouth | Dorset | Solihull | Lancashire | | Southwark |
| Torbay | Buckinghamshire | Walsall | Leicestershire | | Tower Hamlets |
| Swindon | North Northamptonshire | Wolverhampton | Lincolnshire | | Waltham Forest |
| Luton | West Northamptonshire | Bradford | Norfolk | | Wandsworth |
| Southend-on-Sea | Bolton | Calderdale | North Yorkshire | | Westminster |

**Table W5.** List of districts in different vaccination coverage clusters for 2021-22 using three clusters

| High | | | | Medium | | Low |
|---|---|---|---|---|---|---|
| Hartlepool | Southampton | Doncaster | Lancashire | Middlesbrough | Hillingdon | Nottingham |
| Redcar and Cleveland | Isle of Wight | Rotherham | Leicestershire | North Lincolnshire | Hounslow | Peterborough |
| Stockton-on-Tees | County Durham | North Tyneside | Norfolk | Derby | Kingston upon Thames | Manchester |
| Darlington | Cheshire East | South Tyneside | North Yorkshire | Leicester | Sutton | Liverpool |
| Halton | Cheshire West and Chester | Sunderland | Nottinghamshire | Bristol | Kent | Birmingham |
| Warrington | Shropshire | Dudley | Oxfordshire | Luton | Lincolnshire | Barking and Dagenham |
| Blackburn with Darwen | Cornwall | Solihull | Somerset | Thurrock | Surrey | Barnet |
| Blackpool | Wiltshire | Calderdale | Staffordshire | Medway | | Brent |
| Kingston upon Hull | Bedford | Kirklees | Suffolk | Reading | | Camden |
| East Riding of Yorkshire | Central Bedfordshire | Wakefield | Warwickshire | Slough | | Croydon |
| North East Lincolnshire | Northumberland | Gateshead | West Sussex | Brighton and Hove | | Enfield |
| York | Bournemouth, Christchurch and Poole | Cambridgeshire | Worcestershire | Bury | | Greenwich |
| Herefordshire | Dorset | Cumbria | Rutland | Salford | | Hackney |
| Telford and Wrekin | Buckinghamshire | Derbyshire | | Knowsley | | Hammersmith and Fulham |
| Stoke-on-Trent | North Northamptonshire | Devon | | Sefton | | Haringey |
| Bath and North East Somerset | West Northamptonshire | East Sussex | | Sheffield | | Islington |
| North Somerset | Bolton | Essex | | Newcastle upon Tyne | | Kensington and Chelsea |
| South Gloucestershire | Oldham | Gloucestershire | | Coventry | | Lambeth |
| Plymouth | Rochdale | Hampshire | | Sandwell | | Lewisham |
| Torbay | Stockport | Hertfordshire | | Walsall | | Merton |
| Swindon | Tameside | | | Wolverhampton | | Newham |
| Southend-on-Sea | Trafford | | | Bradford | | Redbridge |
| Bracknell Forest | Wigan | | | Leeds | | Richmond upon Thames |
| West Berkshire | St Helens | | | Bexley | | Southwark |
| Windsor and Maidenhead | Wirral | | | Bromley | | Tower Hamlets |
| Wokingham | Barnsley | | | Ealing | | Waltham Forest |
| Milton Keynes | | | | Harrow | | Wandsworth |
| Portsmouth | | | | Havering | | Westminster |

**Table W6.** List of districts in different vaccination coverage clusters for 2021-22 using six clusters

| Highest | | High | | Medium | | Low | Very Low | Lowest |
|---|---|---|---|---|---|---|---|---|
| Redcar and Cleveland | Hampshire | Hartlepool | Rochdale | Middlesbrough | Hillingdon | Nottingham | Barking and Dagenham | Hackney |
| Stockton-on-Tees | Leicestershire | Darlington | Tameside | North Lincolnshire | Hounslow | Peterborough | Barnet | |
| East Riding of Yorkshire | Staffordshire | Halton | Trafford | Derby | Kingston upon Thames | Manchester | Camden | |
| North East Lincolnshire | Worcestershire | Warrington | Wigan | Leicester | Sutton | Liverpool | Croydon | |
| Bath and North East Somerset | Rutland | Blackburn with Darwen | St Helens | Bristol | Kent | Birmingham | Enfield | |
| North Somerset | | Blackpool | Wirral | Luton | Lincolnshire | Brent | Haringey | |
| South Gloucestershire | | Kingston upon Hull | Doncaster | Thurrock | Surrey | Greenwich | Islington | |
| Plymouth | | York | Dudley | Medway | | Hammersmith and Fulham | Kensington and Chelsea | |
| Torbay | | Herefordshire | Solihull | Reading | | Lambeth | Newham | |
| Bracknell Forest | | Telford and Wrekin | Calderdale | Slough | | Lewisham | Redbridge | |
| West Berkshire | | Stoke-on-Trent | Kirklees | Brighton and Hove | | Merton | Waltham Forest | |
| Wokingham | | Swindon | Gateshead | Bury | | Richmond upon Thames | Westminster | |
| Portsmouth | | Southend-on-Sea | Cambridgeshire | Salford | | Southwark | | |
| County Durham | | Windsor and Maidenhead | East Sussex | Knowsley | | Tower Hamlets | | |
| Shropshire | | Milton Keynes | Essex | Sefton | | Wandsworth | | |
| Wiltshire | | Southampton | Gloucestershire | Sheffield | | | | |
| Northumberland | | Isle of Wight | Hertfordshire | Newcastle upon Tyne | | | | |
| Dorset | | Cheshire East | Lancashire | Coventry | | | | |
| Stockport | | Cheshire West and Chester | Norfolk | Sandwell | | | | |
| Barnsley | | Cornwall | North Yorkshire | Walsall | | | | |
| Rotherham | | Bedford | Nottinghamshire | Wolverhampton | | | | |
| North Tyneside | | Central Bedfordshire | Oxfordshire | Bradford | | | | |
| South Tyneside | | Bournemouth, Christchurch and Poole | Somerset | Leeds | | | | |
| Sunderland | | Buckinghamshire | Suffolk | Bexley | | | | |
| Wakefield | | North Northamptonshire | Warwickshire | Bromley | | | | |
| Cumbria | | West Northamptonshire | West Sussex | Ealing | | | | |
| Derbyshire | | Bolton | | Harrow | | | | |
| Devon | | Oldham | | Havering | | | | |

**Table W7**. List of districts in different vaccination coverage clusters for 2022-23 using two clusters

| | High | | | | Low |
|---|---|---|---|---|---|
| Hartlepool | Luton | Buckinghamshire | Sandwell | Gloucestershire | Nottingham |
| Middlesbrough | Southend-on-Sea | North Northamptonshire | Solihull | Hampshire | Peterborough |
| Redcar and Cleveland | Thurrock | West Northamptonshire | Walsall | Hertfordshire | Manchester |
| Stockton-on-Tees | Medway | Bolton | Wolverhampton | Kent | Liverpool |
| Darlington | Bracknell Forest | Bury | Bradford | Lancashire | Birmingham |
| Halton | West Berkshire | Oldham | Calderdale | Leicestershire | Barking and Dagenham |
| Warrington | Reading | Rochdale | Kirklees | Lincolnshire | Barnet |
| Blackburn with Darwen | Slough | Salford | Leeds | Norfolk | Brent |
| Blackpool | Windsor and Maidenhead | Stockport | Wakefield | North Yorkshire | Camden |
| Kingston upon Hull, City of | Wokingham | Tameside | Gateshead | Nottinghamshire | Croydon |
| East Riding of Yorkshire | Milton Keynes | Trafford | Bexley | Oxfordshire | Enfield |
| North East Lincolnshire | Brighton and Hove | Wigan | Bromley | Somerset | Hackney |
| North Lincolnshire | Portsmouth | Knowsley | Ealing | Staffordshire | Hammersmith and Fulham |
| York | Southampton | St. Helens | Greenwich | Suffolk | Haringey |
| Derby | Isle of Wight | Sefton | Harrow | Surrey | Islington |
| Leicester | County Durham | Wirral | Havering | Warwickshire | Kensington and Chelsea |
| Herefordshire, County of | Cheshire East | Barnsley | Hillingdon | West Sussex | Lambeth |
| Telford and Wrekin | Cheshire West and Chester | Doncaster | Hounslow | Worcestershire | Lewisham |
| Stoke-on-Trent | Shropshire | Rotherham | Kingston upon Thames | Rutland | Merton |
| Bath and North East Somerset | Cornwall | Sheffield | Sutton | | Newham |
| Bristol, City of | Wiltshire | Newcastle upon Tyne | Cambridgeshire | | Redbridge |
| North Somerset | Bedford | North Tyneside | Cumbria | | Richmond upon Thames |
| South Gloucestershire | Central Bedfordshire | South Tyneside | Derbyshire | | Tower Hamlets |
| Plymouth | Northumberland | Sunderland | Devon | | Waltham Forest |
| Torbay | Bournemouth, Christchurch and Poole | Coventry | East Sussex | | Wandsworth |
| Swindon | Dorset | Dudley | Essex | | Westminster |

**Table W8**. List of districts in different vaccination coverage clusters for 2022-23 using three clusters

| High | | | Medium | | Low |
|---|---|---|---|---|---|
| Hartlepool | Portsmouth | Dudley | Middlesbrough | Doncaster | Nottingham |
| Redcar and Cleveland | Isle of Wight | Solihull | Blackburn with Darwen | Sheffield | Peterborough |
| Stockton-on-Tees | County Durham | Kirklees | Kingston upon Hull, City of | Newcastle upon Tyne | Manchester |
| Darlington | Cheshire East | Wakefield | North Lincolnshire | Coventry | Liverpool |
| Halton | Cheshire West and Chester | Gateshead | Derby | Sandwell | Birmingham |
| Warrington | Shropshire | Cambridgeshire | Leicester | Walsall | Barking and Dagenham |
| Blackpool | Cornwall | Cumbria | Bristol, City of | Wolverhampton | Barnet |
| East Riding of Yorkshire | Wiltshire | Derbyshire | Luton | Bradford | Camden |
| North East Lincolnshire | Bedford | Devon | Southend-on-Sea | Calderdale | Croydon |
| York | Central Bedfordshire | Essex | Thurrock | Leeds | Enfield |
| Herefordshire, County of | Northumberland | Gloucestershire | Medway | Bexley | Hackney |
| Telford and Wrekin | Bournemouth, Christchurch and Poole | Hampshire | Reading | Brent | Hammersmith and Fulham |
| Stoke-on-Trent | Dorset | Hertfordshire | Slough | Bromley | Haringey |
| Bath and North East Somerset | Buckinghamshire | Lancashire | Milton Keynes | Ealing | Islington |
| North Somerset | Bolton | Leicestershire | Brighton and Hove | Greenwich | Kensington and Chelsea |
| South Gloucestershire | Stockport | Norfolk | Southampton | Harrow | Merton |
| Plymouth | Trafford | North Yorkshire | North Northamptonshire | Havering | Newham |
| Torbay | Wigan | Nottinghamshire | West Northamptonshire | Hillingdon | Redbridge |
| Swindon | St. Helens | Oxfordshire | Bury | Hounslow | Richmond upon Thames |
| Bracknell Forest | Wirral | Somerset | Oldham | Kingston upon Thames | Tower Hamlets |
| West Berkshire | Barnsley | Staffordshire | Rochdale | Lambeth | Waltham Forest |
| Windsor and Maidenhead | Rotherham | Suffolk | Salford | Lewisham | Wandsworth |
| Wokingham | North Tyneside | Warwickshire | Tameside | Southwark | Westminster |
| Rutland | South Tyneside | West Sussex | Knowsley | Sutton | |
| | Sunderland | Worcestershire | Sefton | East Sussex | |
| | | | | Kent | |
| | | | | Lincolnshire | |
| | | | | Surrey | |

**Table W9**. List of districts in different vaccination coverage clusters for 2022-23 using six clusters

| Highest | High | Medium | | Low | Very Low | Lowest |
|---|---|---|---|---|---|---|
| Stockton-on-Tees | Hartlepool | Trafford | Blackburn with Darwen | Calderdale | Middlesbrough | Nottingham | Hackney |
| East Riding of Yorkshire | Redcar and Cleveland | Wigan | Kingston upon Hull, City of | Leeds | Derby | Peterborough | |
| North East Lincolnshire | Darlington | St. Helens | North Lincolnshire | Bromley | Leicester | Manchester | |
| Bath and North East Somerset | Halton | Wirral | Bristol, City of | East Sussex | Luton | Liverpool | |
| North Somerset | Warrington | Rotherham | Southend-on-Sea | Kent | Oldham | Birmingham | |
| South Gloucestershire | Blackpool | Dudley | Thurrock | Lincolnshire | Salford | Barking and Dagenham | |
| Plymouth | York | Solihull | Medway | Surrey | Knowsley | Barnet | |
| West Berkshire | Herefordshire, County of | Kirklees | Reading | | Coventry | Camden | |
| County Durham | Telford and Wrekin | Wakefield | Slough | | Sandwell | Croydon | |
| Shropshire | Stoke-on-Trent | Gateshead | Milton Keynes | | Wolverhampton | Enfield | |
| Wiltshire | Torbay | Cambridgeshire | Brighton and Hove | | Bexley | Hammersmith and Fulham | |
| Northumberland | Swindon | Essex | Southampton | | Brent | Haringey | |
| Dorset | Bracknell Forest | Gloucestershire | North Northamptonshire | | Ealing | Islington | |
| Stockport | Windsor and Maidenhead | Hertfordshire | West Northamptonshire | | Greenwich | Kensington and Chelsea | |
| Barnsley | Wokingham | Lancashire | Bury | | Harrow | Merton | |
| North Tyneside | Portsmouth | Norfolk | Rochdale | | Havering | Newham | |
| South Tyneside | Isle of Wight | North Yorkshire | Tameside | | Hillingdon | Redbridge | |
| Sunderland | Cheshire East | Nottinghamshire | Sefton | | Hounslow | Richmond upon Thames | |
| Cumbria | Cheshire West and Chester | Oxfordshire | Doncaster | | Kingston upon Thames | Tower Hamlets | |
| Derbyshire | Cornwall | Somerset | Sheffield | | Lambeth | Waltham Forest | |
| Devon | Bedford | Suffolk | Newcastle upon Tyne | | Lewisham | Wandsworth | |
| Hampshire | Central Bedfordshire | Warwickshire | Walsall | | Southwark | Westminster | |
| Leicestershire | Bournemouth, Christchurch and Poole | West Sussex | Bradford | | Sutton | | |
| Staffordshire | Buckinghamshire | | | | | | |
| Worcestershire | Bolton | | | | | | |
| Rutland | | | | | | | |

**Table W10**. List of districts in different vaccination coverage clusters for 2023-24 using two clusters

| High | | | | | Low |
|---|---|---|---|---|---|
| Hartlepool | Thurrock | Bolton | Bradford | Lancashire | Nottingham |
| Middlesbrough | Medway | Bury | Calderdale | Leicestershire | Peterborough |
| Redcar and Cleveland | Bracknell Forest | Manchester | Kirklees | Lincolnshire | Luton |
| Stockton-on-Tees | West Berkshire | Oldham | Leeds | Norfolk | Rochdale |
| Darlington | Reading | Salford | Wakefield | North Yorkshire | Sefton |
| Halton | Slough | Stockport | Gateshead | Nottinghamshire | Dudley |
| Warrington | Windsor and Maidenhead | Tameside | Barking and Dagenham | Oxfordshire | Bexley |
| Blackburn with Darwen | Wokingham | Trafford | Barnet | Somerset | Brent |
| Blackpool | Milton Keynes | Wigan | Bromley | Staffordshire | Camden |
| Kingston upon Hull | Brighton and Hove | Knowsley | Croydon | Suffolk | Ealing |
| East Riding of Yorkshire | Portsmouth | Liverpool | Greenwich | Surrey | Enfield |
| North East Lincolnshire | Southampton | St Helens | Hammersmith and Fulham | Warwickshire | Hackney |
| North Lincolnshire | Isle of Wight | Wirral | Hounslow | West Sussex | Haringey |
| York | County Durham | Barnsley | Islington | Worcestershire | Harrow |
| Derby | Cheshire East | Doncaster | Lewisham | Rutland | Havering |
| Leicester | Cheshire West and Chester | Rotherham | Merton | | Hillingdon |
| Herefordshire | Shropshire | Sheffield | Newham | | Kensington and Chelsea |
| Telford and Wrekin | Cornwall | Newcastle upon Tyne | Tower Hamlets | | Kingston upon Thames |
| Stoke-on-Trent | Wiltshire | North Tyneside | Waltham Forest | | Lambeth |
| Bath and North East Somerset | Bedford | South Tyneside | Derbyshire | | Redbridge |
| Bristol | Central Bedfordshire | Sunderland | Devon | | Richmond upon Thames |
| North Somerset | Northumberland | Birmingham | East Sussex | | Southwark |
| South Gloucestershire | Bournemouth, Christchurch and Poole | Coventry | Essex | | Sutton |
| Plymouth | Dorset | Sandwell | Gloucestershire | | Wandsworth |
| Torbay | Buckinghamshire | Solihull | Hampshire | | Westminster |
| Swindon | North Northamptonshire | Walsall | Hertfordshire | | Cambridgeshire |
| Southend-on-Sea | West Northamptonshire | Wolverhampton | Kent | | Cumbria |

**Table W11.** List of districts in different vaccination coverage clusters for 2023-24 using three clusters

| High | | | Medium | | | Low |
|---|---|---|---|---|---|---|
| Hartlepool | Shropshire | Hampshire | Middlesbrough | Wirral | Islington | Nottingham |
| Redcar and Cleveland | Cornwall | Hertfordshire | Blackburn with Darwen | Barnsley | Lewisham | Peterborough |
| Stockton-on-Tees | Wiltshire | Kent | Blackpool | Sheffield | Merton | Luton |
| Darlington | Central Bedfordshire | Norfolk | Kingston upon Hull | North Tyneside | Newham | Rochdale |
| Halton | Northumberland | Nottinghamshire | North Lincolnshire | South Tyneside | Tower Hamlets | Sefton |
| Warrington | Dorset | Oxfordshire | Derby | Sandwell | Waltham Forest | Dudley |
| East Riding of Yorkshire | Buckinghamshire | Somerset | Leicester | Walsall | Gloucestershire | Bexley |
| North East Lincolnshire | Bolton | Staffordshire | Stoke-on-Trent | Wolverhampton | Lancashire | Brent |
| York | Bury | Suffolk | Bristol, City of | Bradford | Leicestershire | Camden |
| Herefordshire | Trafford | Warwickshire | Southend-on-Sea | Calderdale | Lincolnshire | Ealing |
| Telford and Wrekin | Knowsley | West Sussex | Thurrock | Kirklees | North Yorkshire | Enfield |
| Bath and North East Somerset | Liverpool | Worcestershire | Medway | Leeds | Surrey | Hackney |
| North Somerset | Doncaster | Rutland | Reading | Wakefield | | Haringey |
| South Gloucestershire | Rotherham | | Slough | Gateshead | | Harrow |
| Plymouth | Newcastle upon Tyne | | Milton Keynes | Bromley | | Havering |
| Torbay | Sunderland | | Brighton and Hove | Croydon | | Hillingdon |
| Swindon | Birmingham | | Bedford | Greenwich | | Kensington and Chelsea |
| Bracknell Forest | Coventry | | Bournemouth, Christchurch and Poole | Hammersmith and Fulham | | Kingston upon Thames |
| West Berkshire | Solihull | | North Northamptonshire | Hounslow | | Lambeth |
| Windsor and Maidenhead | Barking and Dagenham | | West Northamptonshire | | | Redbridge |
| Wokingham | Barnet | | Manchester | | | Richmond upon Thames |
| Portsmouth | Derbyshire | | Oldham | | | Southwark |
| Southampton | Devon | | Salford | | | Sutton |
| Isle of Wight | East Sussex | | Stockport | | | Wandsworth |
| County Durham | Essex | | Tameside | | | Westminster |
| Cheshire East | | | Wigan | | | Cambridgeshire |
| Cheshire West and Chester | | | St Helens | | | Cumbria |

**Table W12**. List of districts in different vaccination coverage clusters for 2023-24 using six clusters

| Highest | | | High | | Medium | Low | Very Low | Lowest |
|---|---|---|---|---|---|---|---|---|
| Hartlepool | Cornwall | Nottinghamshire | Blackpool | North Yorkshire | Middlesbrough | Nottingham | Bexley | Haringey |
| Redcar and Cleveland | Wiltshire | Oxfordshire | Kingston upon Hull | Surrey | Blackburn with Darwen | Peterborough | Ealing | Lambeth |
| Stockton-on-Tees | Central Bedfordshire | Somerset | North Lincolnshire | | Leicester | Luton | Enfield | |
| Darlington | Northumberland | Staffordshire | Derby | | Thurrock | Rochdale | Hackney | |
| Halton | Dorset | Suffolk | Stoke-on-Trent | | Medway | Sefton | Harrow | |
| Warrington | Buckinghamshire | Warwickshire | Bristol | | Slough | Dudley | Havering | |
| East Riding of Yorkshire | Bolton | West Sussex | Southend-on-Sea | | West Northamptonshire | Brent | Kingston upon Thames | |
| North East Lincolnshire | Bury | Worcestershire | Reading | | Oldham | Camden | Richmond upon Thames | |
| York | Trafford | Rutland | Milton Keynes | | Salford | Hillingdon | Cumbria | |
| Herefordshire | Knowsley | | Brighton and Hove | | Stockport | Kensington and Chelsea | | |
| Telford and Wrekin | Liverpool | | Bedford | | Tameside | Redbridge | | |
| Bath and North East Somerset | Doncaster | | Bournemouth, Christchurch and Poole | | Wigan | Southwark | | |
| North Somerset | Rotherham | | North Northamptonshire | | St Helens | Sutton | | |
| South Gloucestershire | Newcastle upon Tyne | | Manchester | | Barnsley | Wandsworth | | |
| Plymouth | Sunderland | | Wirral | | Sandwell | Westminster | | |
| Torbay | Birmingham | | Sheffield | | Walsall | Cambridgeshire | | |
| Swindon | Coventry | | North Tyneside | | Calderdale | | | |
| Bracknell Forest | Solihull | | South Tyneside | | Kirklees | | | |
| West Berkshire | Barking and Dagenham | | Wolverhampton | | Gateshead | | | |
| Windsor and Maidenhead | Barnet | | Bradford | | Bromley | | | |
| Wokingham | Derbyshire | | Leeds | | Greenwich | | | |
| Portsmouth | Devon | | Wakefield | | Hammersmith and Fulham | | | |
| Southampton | East Sussex | | Croydon | | Hounslow | | | |
| Isle of Wight | Essex | | Waltham Forest | | Islington | | | |
| County Durham | Hampshire | | Gloucestershire | | Lewisham | | | |
| Cheshire East | Hertfordshire | | Lancashire | | Merton | | | |
| Cheshire West and Chester | Kent | | Leicestershire | | Newham | | | |
| Shropshire | Norfolk | | Lincolnshire | | Tower Hamlets | | | |

**Feature Importance for 2022-2023 and 2023-2024**

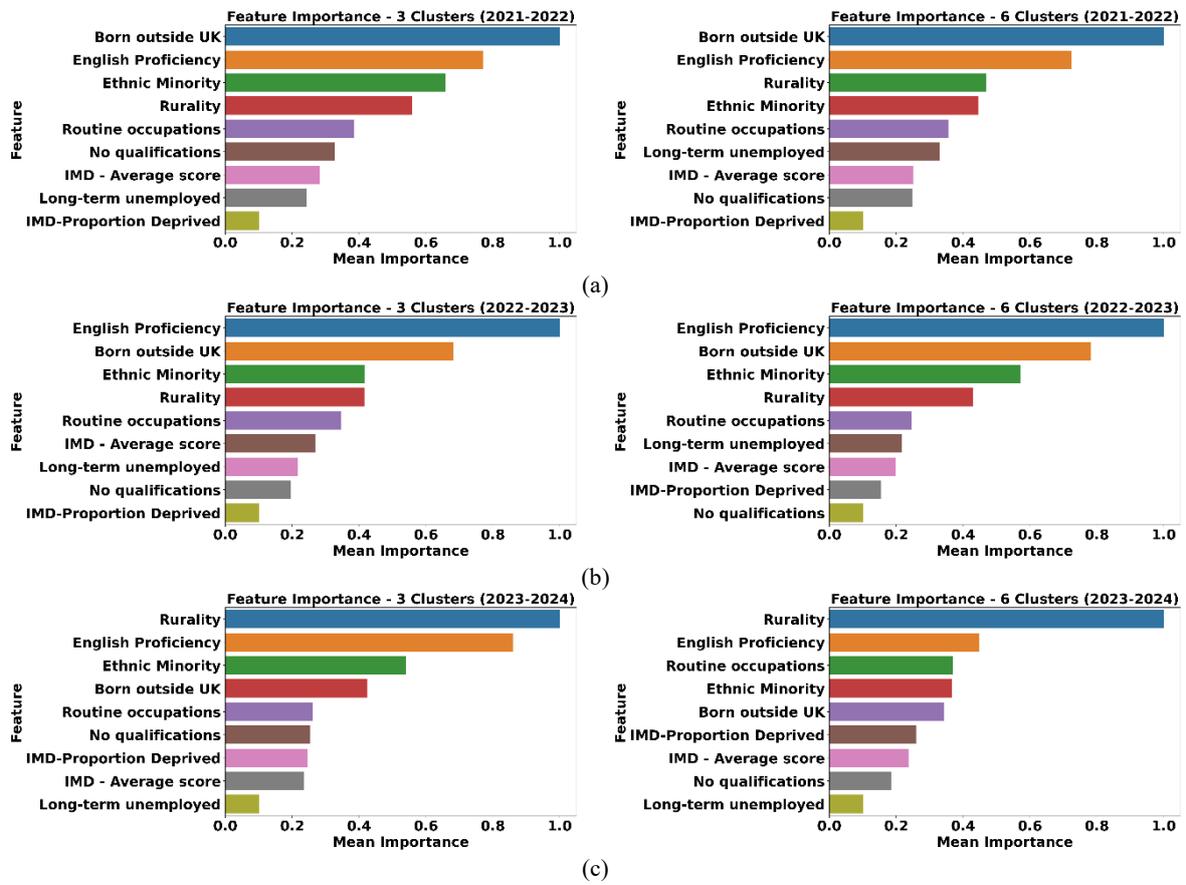

**Figure W4**. The most important determinants of the vaccination coverage rate in England using three and six clusters for (a) 2021-2022, (b) 2022-2023, and (c) 2023-2024.

**Distribution of Rurality Level by Vaccination Cluster**

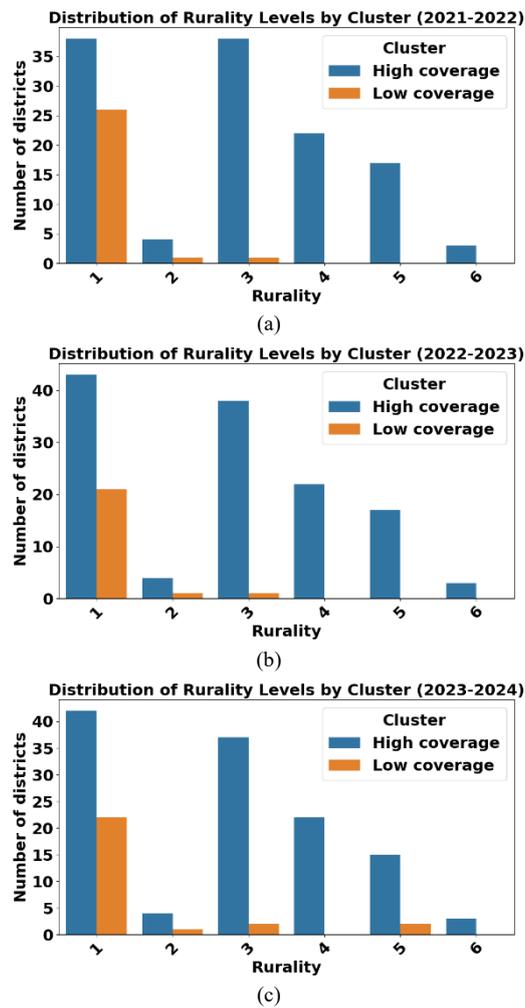

(a)

(b)

(c)

**Figure W5**. Distribution of rurality levels among districts by vaccination coverage cluster for (a) 2021–2022, (b) 2022–2023, and (c) 2023–2024, based on the two-cluster classification.

**Rurality by Vaccination Coverage Cluster**

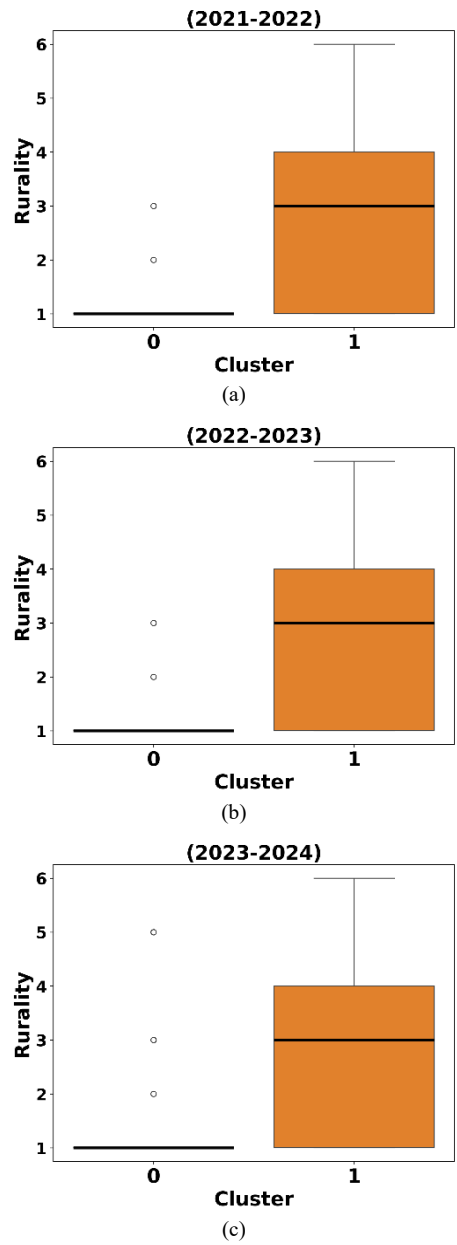

**Figure W6**. Box plot of Rurality by vaccination coverage cluster for (a) 2021-2022, (b) 2022-2023, and (c) 2023-2024, where 0 and 1 show the low and high vaccination coverage clusters, respectively.

**Born outside UK by Vaccination Coverage Cluster**

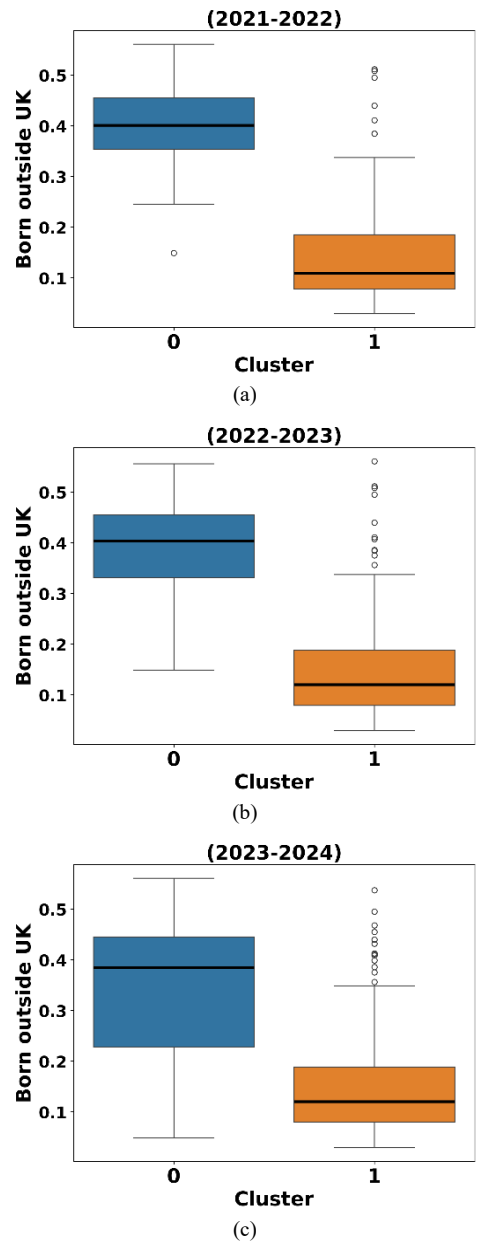

**Figure W7**. Box plot of Born outside UK by vaccination coverage cluster for (a) 2021-2022, (b) 2022-2023, and (c) 2023-2024, where 0 and 1 show the low and high vaccination coverage clusters, respectively.

**English Proficiency by Vaccination Coverage Cluster**

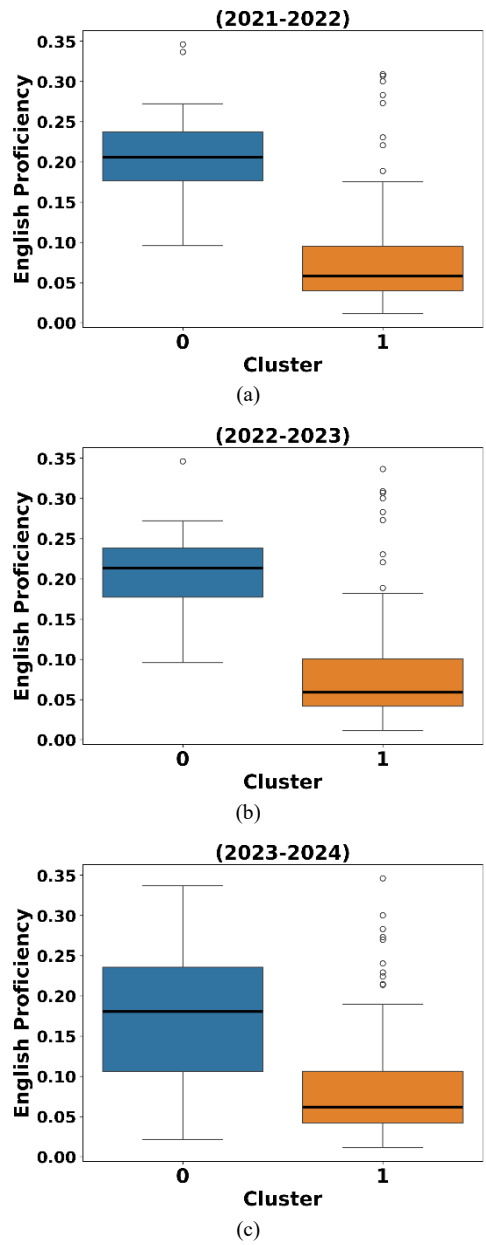

**Figure W8**. Box plot of English Proficiency by vaccination coverage cluster for (a) 2021-2022, (b) 2022-2023, and (c) 2023-2024, where 0 and 1 show the low and high vaccination coverage clusters, respectively.

**Ethnic Minority by Vaccination Coverage Cluster**

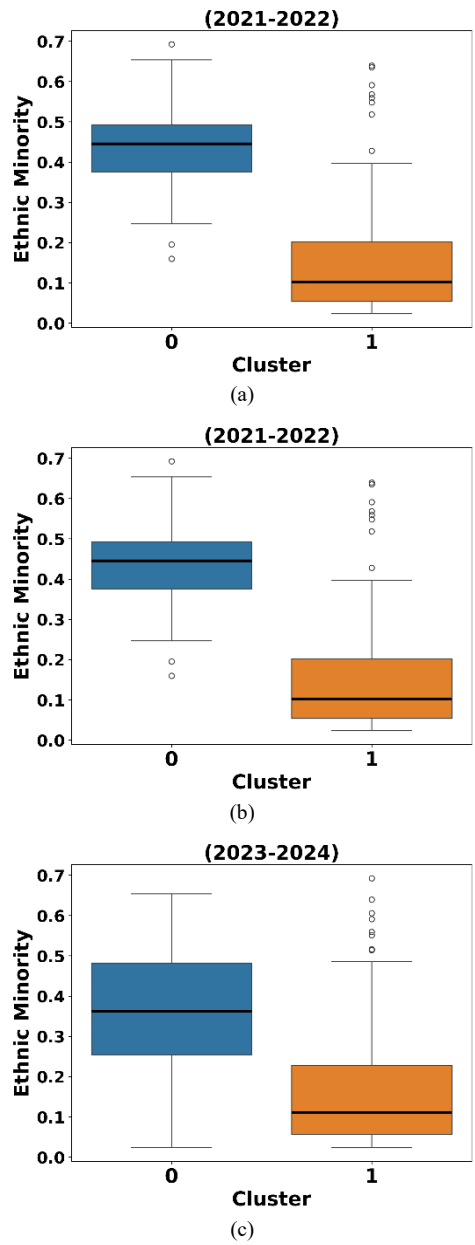

**Figure W9**. Box plot of Ethnic Minority by vaccination coverage cluster for (a) 2021-2022, (b) 2022-2023, and (c) 2023-2024, where 0 and 1 show the low and high vaccination coverage clusters, respectively.